\documentclass[runningheads]{llncs}
\usepackage{graphicx}
\usepackage{tikz}
\usepackage{comment}
\usepackage{amsmath,amssymb} %
\usepackage{color}

\usepackage[accsupp]{axessibility}  %

\usepackage[width=122mm,left=12mm,paperwidth=146mm,height=193mm,top=12mm,paperheight=217mm]{geometry}

\usepackage{multirow}
\usepackage{graphicx}
\usepackage{booktabs}
\usepackage{pifont}
\usepackage{amsmath}
\usepackage{amssymb}
\usepackage{diagbox}
\usepackage{subcaption}
\usepackage{mathrsfs}
\usepackage[noend]{algpseudocode}
\usepackage{algorithmicx,algorithm}
\usepackage{caption}
\usepackage{makecell}
\usepackage{bbm}

\usepackage{colortbl}

\newcommand{\green}[1]{\textcolor[RGB]{96,177,87}{#1}}
\definecolor{mygray}{gray}{.92}
\newcommand{\fn}[1]{\scriptsize{#1}}
\newcommand{\gbf}[1]{\green{\bf{\fn{(#1)}}}}

\usepackage{arydshln}
\usepackage{soul}

\usepackage{orcidlink}

\usepackage{hyperref}

\usepackage[misc]{ifsym}

\hypersetup{hidelinks,
	colorlinks=true,
	allcolors=red,
	pdfstartview=Fit,
	breaklinks=true}

\begin{document}
\pagestyle{headings}
\mainmatter
\def\ECCVSubNumber{4762}  %

\title{AdaBin: Improving Binary Neural Networks with Adaptive Binary Sets} %

\renewcommand\footnotemark{}

\titlerunning{AdaBin: Improving Binary Neural Networks with Adaptive Binary Sets}
\author{Zhijun Tu\inst{1,2}\orcidlink{0000-0001-8740-7927} \and
	Xinghao Chen\inst{2}\thanks{\textsuperscript{\Letter}~Corresponding author}\textsuperscript{(\Letter)}\orcidlink{0000-0002-2102-8235} \and
	Pengju Ren\inst{1}\textsuperscript{(\Letter)}\orcidlink{0000-0003-1163-2014}  \and
	Yunhe Wang\inst{2}\orcidlink{0000-0002-0142-509X}}
\authorrunning{Z. Tu et al.}

\institute{Institute of Artificial Intelligence and Robotics, Xi'an Jiaotong University
\email{tuzhijun123@stu.xjtu.edu.cn}, \email{pengjuren@xjtu.edu.cn} \and
Huawei Noah's Ark Lab\\
\email{\{zhijun.tu, xinghao.chen, yunhe.wang\}@huawei.com}}

\maketitle

\begin{abstract}
This paper studies the Binary Neural Networks (BNNs) in which weights and activations are both binarized into 1-bit values, thus greatly reducing the memory usage and computational complexity. 
Since the modern deep neural networks are of sophisticated design with complex architecture for the accuracy reason, the diversity on distributions of weights and activations is very high. Therefore, the conventional sign function cannot be well used for effectively binarizing full-precision values in BNNs.
To this end, we present a simple yet effective approach called AdaBin to adaptively obtain the optimal binary sets $\{b_1, b_2\}$ ($b_1, b_2\in \mathbb{R}$) of weights and activations for each layer instead of a fixed set (\textit{i.e.}, $\{-1, +1\}$). In this way, the proposed method can better fit different distributions and increase the representation ability of binarized features. 
In practice, we use the center position and distance of 1-bit values to define a new binary quantization function. 
For the weights, we propose an equalization method to align the symmetrical center of binary distribution to real-valued distribution, and minimize the Kullback-Leibler divergence of them. Meanwhile, we introduce a gradient-based optimization method to get these two parameters for activations, which are jointly trained in an end-to-end manner.
Experimental results on benchmark models and datasets demonstrate that the proposed AdaBin is able to achieve state-of-the-art performance. For instance, we obtain a 66.4\% Top-1 accuracy on the ImageNet using ResNet-18 architecture, and a 69.4 mAP on PASCAL VOC using SSD300. %
The PyTorch code is available at \url{https://github.com/huawei-noah/Efficient-Computing/tree/master/BinaryNetworks/AdaBin} and the MindSpore code is available at \url{https://gitee.com/mindspore/models/tree/master/research/cv/AdaBin}.
\end{abstract}

\begin{figure*}
	\centering
	\begin{subtable}[h]{0.505\textwidth}
		\includegraphics[width=1\linewidth]{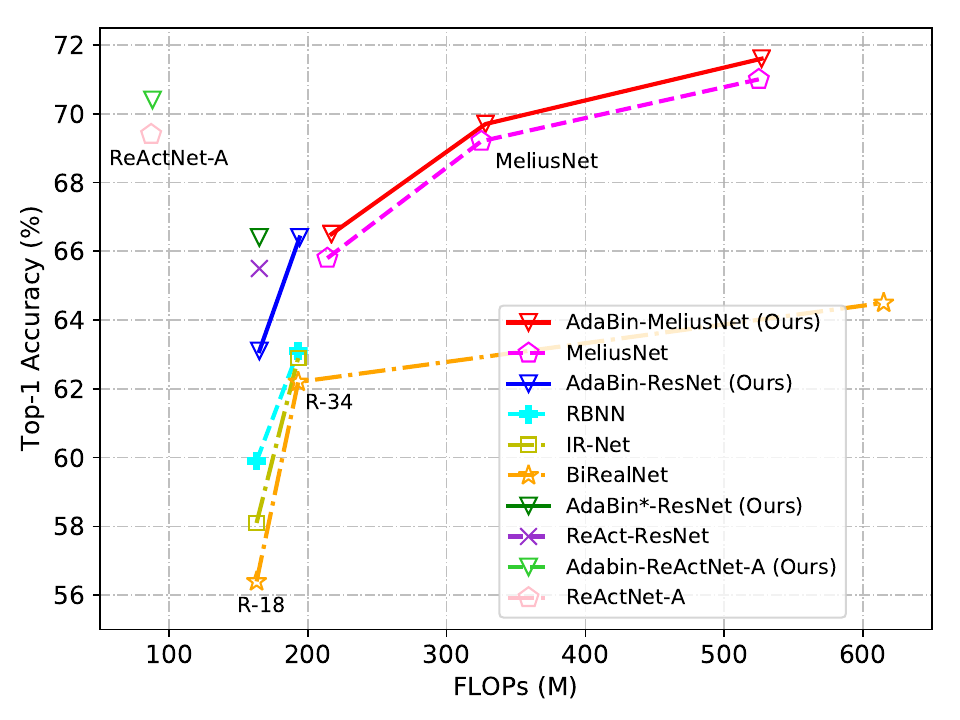}
		\caption{FLOPs vs. ImageNet accuracy}
		\label{Acc_FLOPs}	
	\end{subtable}
	\begin{subtable}[h]{0.445\textwidth}
		\includegraphics[width=1\linewidth]{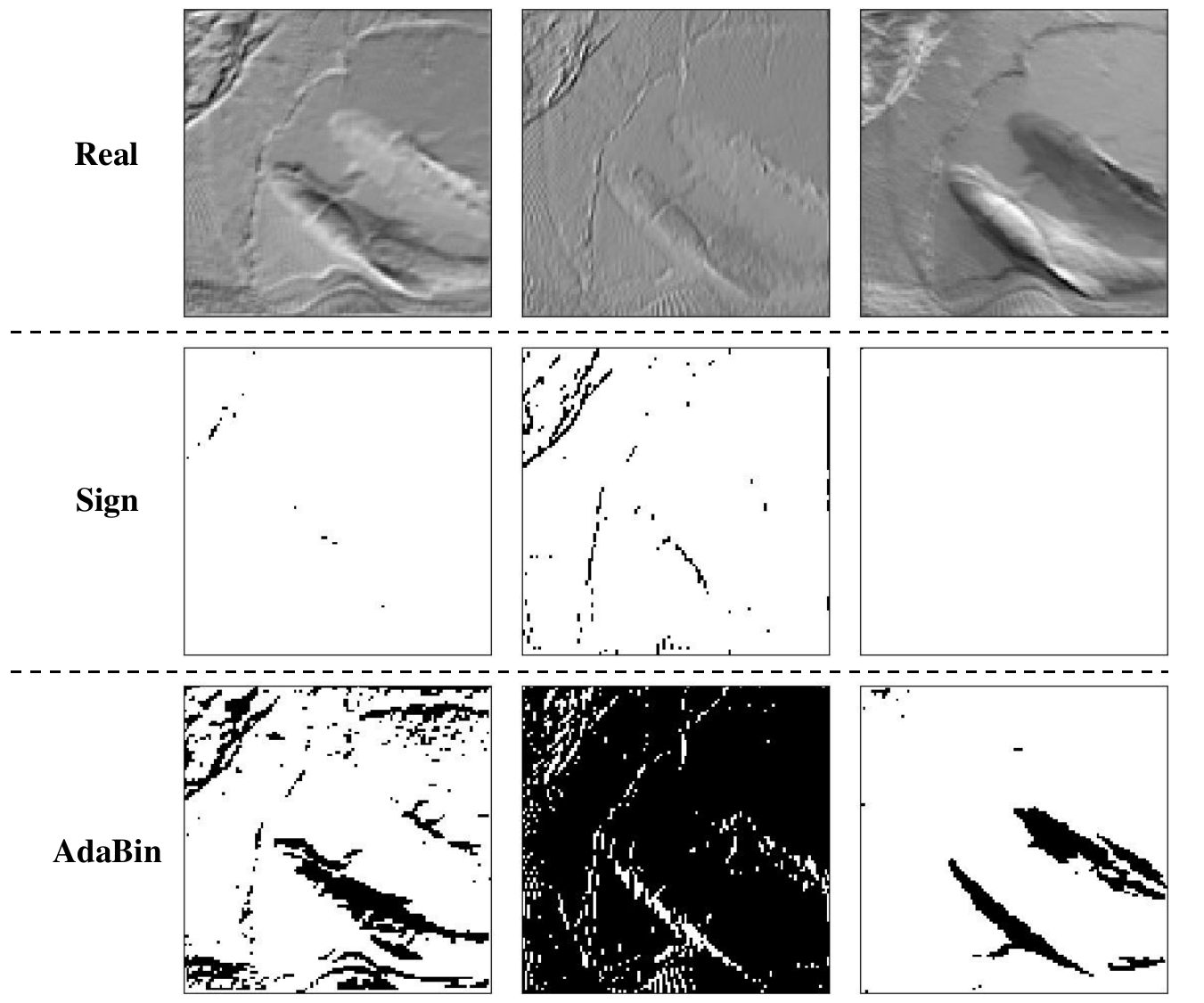}
		\caption{Activations visualization}
		\label{binary_activation}	
	\end{subtable}
	\caption{(a) Comparisons with state-of-the-art methods. With a little extra computation, the proposed AdaBin achieves better results for various architectures such as ResNet, MeliusNet~\cite{bethge2021meliusnet} and ReActNet~\cite{liu2020reactnet}. (b) Visualization for activations of  $2^{nd}$ layer in ResNet-18 on ImageNet. \textit{Real} denotes real-valued activations, \textit{Sign} and \textit{AdaBin} denote the binary methods of previous BNNs and ours.}
\end{figure*}

\section{Introduction}
Deep Neural Networks (DNNs) have demonstrated powerful learning capacity, and are widely applied in various tasks such as computer vision~\cite{krizhevsky2012imagenet}, natural language processing~\cite{bahdanau2014neural} and speech recognition~\cite{hinton2012deep}. 
However, the growing complexity of DNNs requires significant storage and computational resources, which makes the deployment of these deep models on embedded devices extremely difficult. 
Various approaches have been proposed to compress and accelerate DNNs, including low-rank factorization~\cite{yu2017compressing}, pruning~\cite{chen2020mtp,han2015learning}, quantization~\cite{choukroun2019low}, knowledge distillation~\cite{chen2020optical,hinton2015distilling} and energy-efficient architecture design~\cite{chen2020addernet}, \textit{etc}. 
Among these approaches, quantization has attracted great research interests for decades, since the quantized networks with less bit-width require smaller memory footprint, lower energy consumption and shorter calculation delay. 
Binary Neural Networks (BNNs) are the extreme cases of quantized networks and could obtain the largest compression rate by quantizing the weights and activations into 1-bit values~\cite{han2020training,rastegari2016xnor,xu2021learning,yang2020searching}. 
Different from the floating point matrix operation in traditional DNNs, BNNs replace the multiplication and accumulation with bit-wise operation XNOR and BitCount, which can obtain an about $64\times$ acceleration and $32\times$ memory saving~\cite{rastegari2016xnor}. 
However, the main drawback of BNNs is the severe accuracy degradation compared to the full-precision model, which also limits its application to more complex tasks, such as detection, segmentation and tracking.

According to the IEEE-754 standard, a 32-bit floating point number has $6.8\times10^{38}$ unique states~\cite{30711}. In contrast, a 1-bit value only has 2 states $\{b_1, b_2\}$, whose representation ability is very weak compared with that of the full-precision values, since there are only two kinds of the multiplication results of binary values as shown in Table~\ref{BNN}. To achieve a very efficient hardware implementation, the conventional BNN method~\cite{courbariaux2016binarized} binarizes both the weights and the activations to either +1 or -1 with sign function. 
The follow-up approaches on BNNs have made tremendous efforts for enhancing the performance of binary network, 
but still restrict the binary values to a fixed set (\textit{i.e.}, $\{-1, +1\}$ or $\{0, +1\}$) for all the layers.
Given the fact that the feature distributions in deep neural networks are very diverse, sign function can not provide binary diversity for these different distributions. To this end, we have to rethink the restriction of fixed binary set for further enhancing the capacity of BNNs.
\begin{table}[t]
	\center
	\begin{subtable}[h]{0.20\textwidth}
		\setlength{\tabcolsep}{1.5mm}
		\renewcommand\arraystretch{1}
		\centering
		\begin{tabular}{c|cc}
			\hline
			\diagbox{w}{\textit{a}}  & $$-1$$  & $$+1$$ \\ 
			\hline
			$$-1$$ &  $$+1$$  & $$-1$$ \\
			$$+1$$ &  $$-1$$  & $$+1$$ \\
			\hline
		\end{tabular}
		\caption{BNN~\cite{courbariaux2016binarized}}
		\label{BNN}	
	\end{subtable}
	\begin{subtable}[h]{0.20\textwidth}
		\setlength{\tabcolsep}{1.5mm}
		\renewcommand\arraystretch{1}
		\centering
		\begin{tabular}{c|cc}
			\hline
			\diagbox{w}{\textit{a}}  & $$0$$  & $$+1$$ \\ 
			\hline
			$$-1$$ &  $$0$$  & $$-1$$ \\
			$$+1$$ &  $$0$$  & $$+1$$ \\
			\hline
		\end{tabular}
		\caption{SiBNN~\cite{wang2020sparsity}}
		\label{SiBNN}
	\end{subtable}
	\begin{subtable}[h]{0.20\textwidth}
		\setlength{\tabcolsep}{1.5mm}
		\renewcommand\arraystretch{1}
		\centering
		\begin{tabular}{c|cc}
			\hline
			\diagbox{w}{\textit{a}}  & $$-1$$  & $$+1$$ \\ 
			\hline
			$$0$$ &  $$0$$  & $$0$$ \\
			$$+1$$ &  $$-1$$  & $$+1$$ \\
			\hline
		\end{tabular}
		\caption{SiMaN~\cite{lin2021siman}}
		\label{SiMaN}
	\end{subtable}
	\begin{subtable}[h]{0.32\textwidth}
		\setlength{\tabcolsep}{1.5mm}
		\renewcommand\arraystretch{1}
		\centering
		\begin{tabular}{c|cc}
			\hline
			\diagbox{w}{\textit{a}}  & $a_{b1}$  & $a_{b2}$ \\ 
			\hline
			$\textrm{w}_{b1}$ &  $a_{b1}\textrm{w}_{b1}$  & $a_{b2}\textrm{w}_{b1}$ \\
			$\textrm{w}_{b2}$ &  $a_{b1}\textrm{w}_{b2}$  & $a_{b2}\textrm{w}_{b2}$ \\
			\hline
		\end{tabular}
		\caption{AdaBin (Ours)}
		\label{Ours}
	\end{subtable}
	\caption{The illustration on the feature representation ability of different binary schemes. The $a$ represents the binarized input and the $\textrm{w}$ represents binarized weights, respectively. $a_{b1}, a_{b2}, \textrm{w}_{b1}, \textrm{w}_{b2}\in \mathbb{R}$, which are not restricted to fixed values for different layers.}
	\label{truth_table}
\end{table}

Based on the above observation and analysis, we propose an \textit{\textbf{Ada}ptive \textbf{Bin}ary} method (\textit{\textbf{AdaBin}}) to redefine the binary values ($b_1, b_2\in \mathbb{R}$) with their center position and distance, which aims to obtain the optimal binary set that best matches the real-valued distribution. 
We propose two corresponding optimization strategies for weights and activations.
On one hand, we introduce an equalization method for the weights based on statistical analysis. By aligning the symmetrical center of binary distribution to real-valued distribution and minimizing the Kullback-Leibler divergence (KLD) of them, we can obtain the analytic solutions of center and distance, which makes the weight distribution much balanced.
On the other hand, we introduce a gradient-based optimization method for the activations with a loss-aware center and distance, which are initialized in the form of sign function and trained in an end-to-end manner.
As shown in Table~\ref{truth_table}, we present the truth tables of the
multiplication results for binary values in different BNNs. Most previous BNNs binarize both the weights and activations into $\{-1,+1\}$ as shown in Table~\ref{BNN}. A few other methods~\cite{lin2021siman,wang2020sparsity} attempt to binarize weights and activations into $\{0,+1\}$, as shown in Table~\ref{SiBNN} and Table~\ref{SiMaN}. These methods result in 2 or 3 kinds of output representations. Table~\ref{Ours} illustrates the results of our proposed AdaBin method. The activations and weights are not fixed and could provide 4 kinds of output results, which significantly enhances the feature representation of binary networks as shown in Fig.~\ref{binary_activation}. Meanwhile, we can find that previous binary methods are the special cases of our AdaBin and we extend the binary values from $\pm1$ to the whole real number domain.

Furthermore, we demonstrate that the proposed AdaBin can also be efficiently implemented by XNOR and BitCount operations with negligible extra calculations and parameters, which could achieve $60.85\times$ acceleration and $31\times$ memory saving in theory. With only minor extra computation, our proposed AdaBin outperforms state-of-the-art methods for various architectures, as shown in Fig.~\ref{Acc_FLOPs}.
The contributions of this paper are summarize as follow:

(1) We rethink the limitation of $\{-1, +1\}$ in previous BNNs and propose a simple yet effective binary method called AdaBin, which could seek suitable binary sets by adaptively adjusting the center and distance of 1-bit values. 

(2) Two novel strategies are proposed to obtain the optimal binary sets of weights and activations for each layer, which can further close the performance gap between binary neural networks and full-precision networks.

(3) Extensive experiments on CIFAR-10 and ImageNet demonstrate the superior performance of our proposed AdaBin over state-of-the-art methods. Besides, though not tailored for object detection task, AdaBin also outperforms prior task-specific BNN methods by 1.9 mAP on PASCAL VOC dataset.

\section{Related Work}\label{Related_work}
Binary neural network was firstly introduced by~\cite{courbariaux2016binarized}. They creatively proposed to binarize weights and activations with sign function and replace most arithmetic operations of deep neural networks with bit-wise operations.
To reduce the quantization error, XNOR-Net~\cite{rastegari2016xnor} proposed a channel-wise scaling factor to reconstruct the binarized weights, which also becomes one of the most important components of the subsequent BNNs. ABC-Net~\cite{lin2017towards} approximated full-precision weights with the linear combination of multiple binary weight bases and employed multiple binary activations to alleviate information loss. 
Inspired by the structures of ResNet~\cite{he2016deep} and DenseNet~\cite{huang2017densely}, Bi-Real Net~\cite{liu2018bi} proposed to add shortcuts to minimize the performance gap between the 1-bit and real-valued CNN models, and BinaryDenseNet~\cite{bethge2019back}  improved the accuracy of BNNs by increasing the number of concatenate shortcut.
IR-Net~\cite{qin2020forward} proposed the Libra-PB, which can minimize the information loss in forward propagation by maximizing the information entropy of the quantized parameters and minimizing the quantization error with the constraint $\{-1, +1\}$. ReActNet~\cite{liu2020reactnet} proposed to generalize the traditional sign and PReLU functions, denoted as RSign and RPReLU for the respective generalized functions, to enable explicit learning of the distribution reshape and shift at near-zero extra cost. 
\section{Binarization with Adaptive Binary Sets}\label{AdaBin}
In this section, we focus on how to binarize weights and activations respectively, and introduce a new non-linear module to enhance the capacity of BNNs.
\begin{figure}[t]
	\begin{center}
		\includegraphics[width=0.88\linewidth]{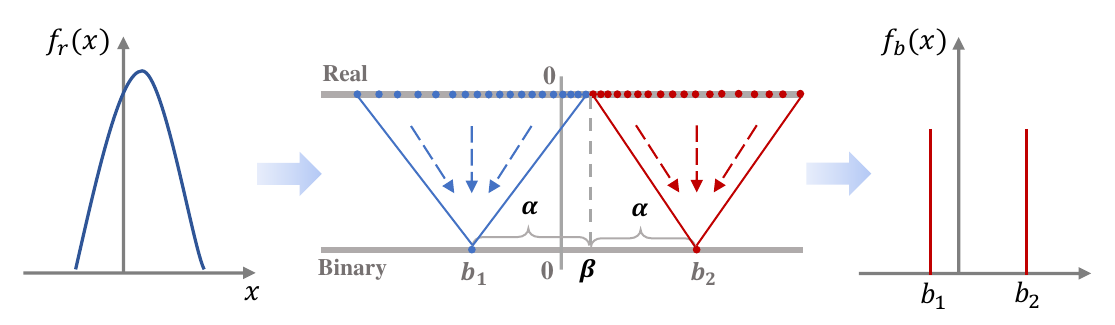}
	\end{center}
	\caption{AdaBin quantizer. The middle represents the mapping from floating point distribution $f_r(x)$ to binary distribution $f_b(x)$. $b_1$ and $b_2$ are the two clusters, $\alpha$ and $\beta$ are the distance and center, respectively.}
	\label{quantizer}
\end{figure}

We first give a brief introduction on the general binary neural networks. 
Given an input $a \in \mathbb{R}^{c\times h\times w}$ and weight $\textrm{w} \in \mathbb{R}^{n\times c\times k\times k}$, then we can get the output $y \in \mathbb{R}^{n\times h'\times w'}$ by convolution operation as Eq.~\ref{equ1}.
\begin{equation}
y = \textrm{Conv}(a, \textrm{w}).
\label{equ1}
\end{equation}
To accelerate the inference process, previous BNNs always partition the input and weight into two clusters, $-1$ and $+1$ with sign function as Eq.~\ref{equ2}.
\begin{equation}
\textrm{Sign}(x)=\left\{
\begin{aligned}
b_1 = -1, & \quad x<0 \\
b_2 = +1, & \quad x\ge0 \\
\end{aligned}
.\right.
\label{equ2}
\end{equation}
Then the floating-point multiplication and accumulation could be replaced by bit-wise operation XNOR (denoted as $\odot$) and BitCount as Eq.~\ref{equ3}, which will result in much less overhead and latency.
\begin{equation}
y = \textrm{BitCount}(a_b \odot \textrm{w}_b).
\label{equ3}
\end{equation}
In our method, we do not constrain the binarized values to a fixed set like $\{-1,+1\}$. Instead we release $b_1$ and $b_2$ to the whole real number domain and utilize the proposed AdaBin quantizer, which could adjust the center position and distance of the two clusters adaptively as Eq.~\ref{equ26}. In this way,  the binarized distribution can best match the real-valued distribution:
\begin{equation}
\mathcal{B}(x) =\left\{
\begin{aligned}
b_1 = \beta-\alpha, & \quad x<\beta \\
b_2 = \beta+\alpha, & \quad x\ge\beta \\
\end{aligned}
,\right.
\label{equ26}
\end{equation}
where the $\alpha$ and $\beta$ are the half-distance and center of the binary values $b_1$ and $b_2$. Fig.~\ref{quantizer} shows the binarization of AdaBin, as we can see that, the data on the left of the center will be clustered into $b_1$ and the data on the right of the center will be clustered into $b_2$. The distance $\alpha$ and center $\beta$ will change with different distributions, which help partition the floating point data into two optimal clusters adaptively. 
For the binarization of weights and activations, we exploit the same form of AdaBin  but different optimization strategies.
\subsection{Weight Equalization}
Low-bit quantization greatly weaken the feature extraction ability of filter weights, especially for 1-bit case. Previous BNNs exploit different methods to optimize the binarized weights. 
XNOR-Net~\cite{rastegari2016xnor} minimizes the mean squared error (MSE) by multiplying a scale factor, and IR-Net~\cite{qin2020forward} obtains the maximum information entropy by weight reshaping and then conduct the same operation as XNOR-Net. However, these methods can not get accurate quantization error between binarized data and real-valued data due to the following limitations. Firstly, the center position of previous binarized values $\{-1, +1\}$ is always $0$, which is not aligned with the center of original real-valued weights. Secondly, MSE is a simple metric to evaluate the quantization error but do not consider the distribution similarity between binarized data and real-valued data. On the contrary, the Kullback-Leibler divergence (KLD) is a measure on probability distributions~\cite{kullback1951information} and is  more accurate to evaluate the information loss than MSE. Therefore, we propose to minimize the KLD to achieve a better distribution-match. We apply the AdaBin for weights binarization as Eq.~\ref{equ6}:
\begin{equation}
\textrm{w}_b = \mathcal{B}(\textrm{w}) =\left\{
\begin{aligned}
\textrm{w}_{b1} = \beta_w-\alpha_w, & \quad \textrm{w}<\beta_w \\
\textrm{w}_{b2} = \beta_w+\alpha_w, & \quad \textrm{w}\ge\beta_w \\
\end{aligned}
,\right.
\label{equ6}
\end{equation}
where $\alpha_w$ and $\beta_w$ are distance and center of binarized weights, the binary elements of $\text{w}_b$ in the forward is $\beta_w-\alpha_w$ and $\beta_w+\alpha_w$. And the KLD of real-valued distribution and binary distribution can be represented as Eq.~\ref{equ7}. 
\begin{equation}
D_{KL}(P_r || P_b) = \int_{x\in \textrm{w}\&\textrm{w}_b } P_r(x)\textrm{log}\frac{P_r(x)}{P_b(x)} dx,
\label{equ7}
\end{equation}
where the $P_r(x)$ and $P_b(x)$ denote the distribution probability of real-valued weights and binarized weights. In order to make the binary distribution more balanced, we need to align its symmetrical center (position of mean value) to the real-valued distribution, so that Eq.~\ref{equ71} can be obtained.
\begin{equation}
\beta_w = \mathbb{E}(\textrm{w}) \approx \frac{1}{c\times k \times k} \sum_{m=0}^{c-1} \sum_{j=0}^{k-1} \sum_{i=0}^{k-1} \textrm{w}_{m,j,i}.
\label{equ71}
\end{equation} 
Therefore, we can further infer that $P_b(\textrm{w}_{b1}) = P_b(\textrm{w}_{b2}) = 0.5$. Since there is no convinced formula of weight distribution for neural networks, it is difficult to calculate the Kullback-Leibler divergence explicitly. However, the weights in such networks typically assume a bell-shaped distribution with tails~\cite{anderson2017high,baskin2021uniq,zhang2021differentiable}, and the both sides are symmetrical on the center,
then we can get the $\alpha_w$ as Eq.~\ref{equ72}, the detailed proof is in the supplementary.
\begin{equation}
\alpha_w = \frac{{\Vert \textrm{w}-\beta_w\Vert_2}}{\sqrt{c\times k \times k}},
\label{equ72}
\end{equation} 
where $\Vert \cdot \Vert_2$ denotes the $\ell_2$-norm. In our method, the distance $\alpha_w$ and center $\beta_w$ are channel-wise parameters for weight binarization, and updated along the real-valued weights during the training process. Without distribution reshaping and the constraint that the center of binary values is $0$ as IR-Net and XNOR-Net, AdaBin could equalize the weights to make the binarized distribution best match the real-valued distribution.
 During the inference, we can decompose the binary weights matrix into 1-bit storage format as following:
\begin{equation}
\textrm{w}_b = \alpha_w b_w + \beta_w, b_w \in \{-1,+1\}.
\label{weight_binarization}
\end{equation}
So that the same as the previous BNNs, our method can also achieve about 32$\times$ memory saving.
\subsection{Gradient-based Activation Binarization}
Activation quantization is a challenging task with low bit-width, and has much more impacts to the final performance than weight. HWGQ~\cite{cai2017deep} proposed to address this challenge by applying a half-wave Gaussian quantization method, based on the observation that activation after Batch Normalization tends to have a symmetric, non-sparse distribution, that is close to Gaussian and ReLU is a half-wave rectifier. However, recent BNNs~\cite{martinez2020training} proposed to replace the ReLU with PReLU~\cite{he2015delving}, which could facilitate the training of binary networks. So that HWGQ can not be further applied because of this limitation. Besides, the distribution of real-valued activations is not as stable as weights, which  keeps changing for different inputs. Therefore we can not extract the center and distance from the activations as Eq.~\ref{equ71} and Eq.~\ref{equ72}, which brings extra cost to calculate them and will greatly weaken the hardware efficiency of binary neural networks during inference. 
In order to get the optimal binary activation during training, we propose a gradient-based optimization method to minimize the accuracy degradation arising from activation binarization. Firstly, we apply the AdaBin quantizer to activations as Eq.~\ref{equ6.2}.
\begin{equation}
a_b = \mathcal{B}(a) =\left\{
\begin{aligned}
a_{b1} = \beta_a-\alpha_a, & \quad a<\beta_a \\
a_{b2} = \beta_a+\alpha_a, & \quad a\ge\beta_a \\
\end{aligned}
,\right.
\label{equ6.2}
\end{equation}
where $\alpha_a$ and $\beta_a$ are the distance and center of binarized activations, and the binary set of $a_b$ in the forward is $\{\beta_a-\alpha_a,  \beta_a+\alpha_a\}$. To make the binary activations adapt to the dataset as much as possible during the training process, we set $\alpha_a$ and $\beta_a$ as learnable variables, which could be optimized via backward gradient propagation as total loss decreases. In order to ensure that the training process can converge, we need to clip out the gradient of large activation values in the backward as Eq.~\ref{equ6.3}.
\begin{equation}
\frac{\partial \mathcal{L}}{\partial a} = \frac{\partial \mathcal{L}}{\partial a_b}*\mathbbm{1}_{\lvert \frac{a-\beta_a}{\alpha_a} \rvert \le 1},
\label{equ6.3}
\end{equation}
where $\mathcal{L}$ denotes the output loss, $a$ is the real-valued activation and $a_b$ is the binarized activation, $\mathbbm{1}_{\lvert x \rvert \le 1}$ denotes the indicator function that equals to $1$ if ${\lvert x \rvert \le 1}$ is true and $0$ otherwise. This functionality can be achieved by a composite function of hard tanh and sign, thus we rewrite the Eq.~\ref{equ6.2} as following:
\begin{equation}
\begin{aligned}
a_b & = \alpha_a \times \textrm{Sign} (\textrm{Htanh} (\frac{a-\beta_a}{\alpha_a})) + \beta_a.
\label{equ6.4}
\end{aligned}
\end{equation}
For simplicity, we denote $g(x)=\textrm{Sign} (\textrm{Htanh}(x))$, then we can get the gradient of $\alpha_a$ and $\beta_a$ as Eq.~\ref{equ10} in the backward:
\begin{equation}\small
\begin{aligned}
\frac{\partial \mathcal{L}}{\partial \alpha_a} &= \frac{\partial \mathcal{L}}{\partial a_b} \frac{\partial a_b}{\partial \alpha_a}= \frac{\partial \mathcal{L}}{\partial a_b}(g(\frac{a-\beta_a}{\alpha_a})-\frac{a}{\alpha_a}g'(\frac{a-\beta_a}{\alpha_a})) ,\\
\frac{\partial \mathcal{L}}{\partial \beta_a} &= \frac{\partial \mathcal{L}}{\partial a_b} \frac{\partial a_b}{\partial \beta_a} = \frac{\partial \mathcal{L}}{\partial a_b}(1-g'(\frac{a-\beta_a}{\alpha_a})),
\end{aligned}
\label{equ10}
\end{equation}
where $g'(x)$ is the derivative of $g(x)$. We set the initial values of center position $\beta_a$ and distance $\alpha_a$ to $0$ and $1$, so that the initial effect of our binary quantizer is equivalent to the sign function\cite{courbariaux2016binarized,liu2018bi,qin2020forward}. Then these two parameters of different layers are dynamically updated via gradient descent-based training, and converge to the optimal center and distance values, which is much different from the unified usage of the sign function in the previous BNNs. During inference, the $\alpha_a$ and $\beta_a$ of all the layers are fixed, then we can binarize the floating point activations into 1-bit as followings:
\begin{equation}
a_b = \alpha_a b_a + \beta_a, b_a \in \{-1,+1\},
\label{equ100}
\end{equation}
where the $b_a$ is the 1-bit storage form and obtained online with input data. Compared with the sign function of previous BNNs, AdaBin will take a little overhead but could significantly improve the feature capacity of activations with the adaptive binary sets for each layer.
\begin{figure}[t]
	\begin{center}
		\includegraphics[height=0.5\linewidth]{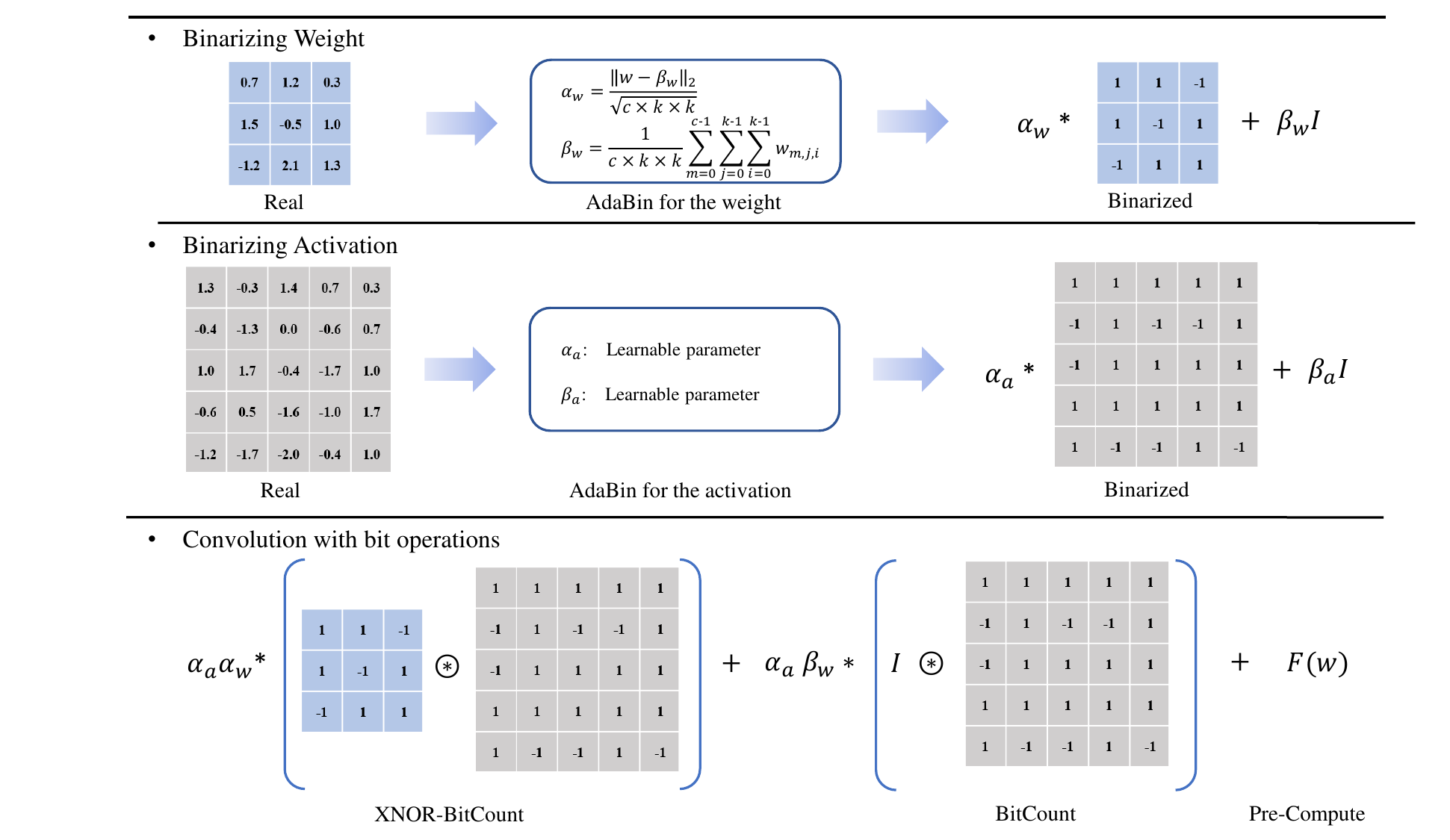}
	\end{center}
	\caption{Binary convolution process. The $I$ represents the identity matrix, and $F(\textrm{w})$ represents the extra computation with $\textrm{w}$, which could be pre-computed during the inference.}
	\label{binary_convolution}
\end{figure}
\subsection{Non-linearity}
Prior methods~\cite{martinez2020training} propose to use Parametric Rectified Linear Unit (PReLU)~\cite{he2015delving} as it is known to facilitate the training of binary networks. PReLU adds an adaptively learnable scaling factor in the negative part and remain unchanged in the positive part. However, we empirically found that the binary values with our proposed AdaBin are almost all positive in very few layers, which invalidate the non-linearity of PReLU. Therefore, to further enhance the representation of feature maps, we propose to utilize Maxout~\cite{goodfellow2013maxout} for the stronger non-linearity in our AdaBin, which is defined as Eq.~\ref{equ11}.
\begin{equation}
f_c(x) = \gamma_c^+ \textrm{ReLU}(x) - \gamma_c^- \textrm{ReLU}(-x),
\label{equ11}  
\end{equation}
where $x$ is the input of the Maxout function, $\gamma_c^+$ and $\gamma_c^-$ are the learnable coefficient for the positive part and negative part of the $c$-th channel, respectively. Following the setting of PReLU, the initialization of $\gamma_c^+$ and $\gamma_c^-$ are 1 and 0.25.
\begin{table}[tb]
	\setlength{\tabcolsep}{3.5mm}
	\renewcommand\arraystretch{1.0}
	\begin{center}
		\begin{tabular}{clcccc}
			\toprule[1pt]
			Networks                   & Methods &     W/A  & Acc. (\%) \\
			\midrule[1pt]
			\multirow{6}{*}{ResNet-18} & Full-precision     &         32/32      & 94.8     \\
			\cline{2-6}
			& RAD~\cite{ding2019regularizing}    &  \multirow{5}{*}{1/1} & 90.5     \\
			& IR-Net~\cite{qin2020forward} &                       & 91.5     \\
			& RBNN~\cite{lin2020rotated}   &                      & 92.2     \\
			& ReCU~\cite{xu2021recu}   &                       & 92.8     \\
			& \textbf{AdaBin (Ours)}   &             & \textbf{93.1}     \\
			\midrule[1pt]
			\multirow{5}{*}{ResNet-20} & Full-precision      &         32/32        & 91.7     \\
			\cline{2-6}
			& DoReFa~\cite{zhou2016dorefa}  &  \multirow{5}{*}{1/1} & 79.3     \\
			& DSQ~\cite{gong2019differentiable}     &                       & 84.1     \\
			& IR-Net~\cite{qin2020forward}  &                       & 86.5     \\
			& RBNN~\cite{lin2020rotated}    &                      & 87.8     \\
			& \textbf{AdaBin (Ours)}    &                      & \textbf{88.2}     \\
			\midrule[1pt]
			\multirow{9}{*}{VGG-Small} & Full-precision      &         32/32        & 94.1     \\
			\cline{2-6}
			& LAB~\cite{hou2016loss}     &  \multirow{8}{*}{1/1}  & 87.7     \\
			& XNOR-Net~\cite{rastegari2016xnor}    &                       & 89.8     \\
			& BNN~\cite{courbariaux2016binarized}     &                      & 89.9     \\
			& RAD~\cite{ding2019regularizing}     &                      & 90.0     \\
			& IR-Net~\cite{qin2020forward}  &                       & 90.4     \\
			& RBNN~\cite{lin2020rotated}    &                       & 91.3     \\
			& SLB\cite{yang2020searching}                  &                       & 92.0 \\
			& \textbf{AdaBin (Ours)}    &                       & \textbf{92.3}     \\
			\bottomrule[1pt]
		\end{tabular}
	\end{center}
	\caption{Comparisons with state-of-the-art methods on CIFAR-10. W/A denotes the bit width of weights and activations.}
	\label{cifar10_results}
\end{table}

\subsection{Binary Convolution for AdaBin}
The goal of BNNs is to replace the computationally expensive multiplication and accumulation with XNOR and BitCount operations. 
Although the binary sets are not limited to $\{-1, +1\}$, our method can still be accelerated with bit-wise operations by simple linear transformation. As shown in Fig.~\ref{binary_convolution}, we can binarize the weights and get the 1-bit matrix $b_w$ offline via Eq.~\ref{weight_binarization}, and binarize the activations to get the 1-bit activations $b_a$ online via Eq.~\ref{equ100}, then decompose the binary convolution into three items. The first term is the same as the previous BNNs, and the second term only needs to accumulation for one output channel, which can be replaced by BitCount. The third term $F(\textrm{w})$ could be pre-computed in the inference process. For $n=c=256$, $k=3$, $w'=h'=14$, compared with the binary convolution of IR-Net~\cite{rastegari2016xnor}, our method only increases 2.74\% operations and 1.37\% parameters, which are negligible compared to the total complexity and could achieve $60.85\times$ acceleration and $31\times$ memory saving in theory, the detailed analysis is shown in the supplementary material.

\section{Experiments}\label{experiment}
In this section, we demonstrate the effectiveness of our proposed AdaBin via comparisons with state-of-the-art methods and extensive ablation experiments. 
We implement the proposed method using PyTroch~\cite{pytorch} and MindSpore~\cite{mindspore}.

\subsection{Results on CIFAR-10}
We train AdaBin for 400 epochs with a batch size of 256, where the initial learning rate is set to 0.1 and then decay with CosineAnnealing as IR-Net\cite{qin2020forward}. We adopt SGD optimizer with a momentum of 0.9, and use the same data augmentation and pre-processing in~\cite{he2016deep} for training and testing. We compare AdaBin with  BNN~\cite{courbariaux2016binarized}, LAB~\cite{hou2016loss}, XNOR-Net~\cite{rastegari2016xnor}, DoReFa~\cite{zhou2016dorefa}, DSQ~\cite{gong2019differentiable}, RAD~\cite{ding2019regularizing}, IR-Net~\cite{qin2020forward}, RBNN~\cite{lin2020rotated}, ReCU~\cite{xu2021recu} and SLB\cite{yang2020searching}.
Table~\ref{cifar10_results} shows the performance of these methods on CIFAR-10. AdaBin obtains 93.1\% accuracy for ResNet-18 architecture, which outperforms the ReCU by 0.3\% and reduces the accuracy gap between BNNs and floating-point model to 1.7\%. Besides, AdaBin obtains 0.4\% accuracy improvement on ResNet-20 compared to the current best method RBNN, and gets 92.3\% accuracy while binarizing the weights and activations of VGG-small into 1-bits, which outperforms SLB by 0.3\%.
\subsection{Results on ImageNet}
We train our proposed AdaBin for 120 epochs from scratch and use SGD optimizer with a momentum of 0.9. We set the initial learning rate to 0.1 and then decay with CosineAnnealing following IR-Net\cite{qin2020forward}, and utilize the same data augmentation and pre-processing in~\cite{he2016deep}. In order to demonstrate the generality of our method, we conduct experiments on two kinds of structures. The first group is the common architectures that are widely used in various computer vision tasks, such as AlexNet~\cite{krizhevsky2012imagenet} and ResNet~\cite{he2016deep}. Another kind is the binary-specific structures such as BDenseNet~\cite{bethge2019binarydensenet}, MeliusNet~\cite{bethge2021meliusnet} and ReActNet~\cite{liu2020reactnet}, which are designed for BNNs and could significantly improve the accuracy with the same amount of parameters as common structures. 

\begin{table}[H]
	\setlength{\tabcolsep}{2mm}
	\renewcommand\arraystretch{0.91}
	\begin{center}
		\begin{tabular}{clccc}
			\toprule[1pt]
			Networks                   & Methods &     W/A  & Top-1 (\%) & Top-5 (\%) \\
			\midrule[1pt]
			\multirow{6}{*}{AlexNet} & Full-precision     &         32/32         & 56.6  & 80.0     \\
			\cline{2-5}
			& BNN~\cite{courbariaux2016binarized}                  & \multirow{5}{*}{1/1}  & 27.9  & 50.4     \\
			& DoReFa~\cite{zhou2016dorefa}               &                       & 43.6  & -  \\
			& XNOR~\cite{rastegari2016xnor}                 &                       & 44.2  & 69.2  \\
			& SiBNN~\cite{wang2020sparsity}                &                       & 50.5  & 74.6  \\
			& \textbf{AdaBin (Ours)}        &                     &  \textbf{53.9}   &   \textbf{77.6}   \\
			\midrule[1pt]
			\multirow{14}{*}{ResNet-18} & Full-precision     &         32/32        & 69.6  & 89.2     \\
			\cline{2-5}
			& BNN~\cite{courbariaux2016binarized}                     & \multirow{7}{*}{1/1}  & 42.2  & -     \\
			& XNOR-Net~\cite{rastegari2016xnor}                     &                       & 51.2  & 73.2  \\
			& Bi-Real~\cite{liu2018bi}                  &                       & 56.4  & 79.5  \\
			& IR-Net~\cite{qin2020forward}                   &                      & 58.1  & 80.0  \\
			& Si-BNN~\cite{wang2020sparsity}                    &                     & 59.7  & 81.8  \\
			& RBNN~\cite{lin2020rotated}                     &                    & 59.9  & 81.9  \\
			& SiMaN~\cite{lin2021siman}                   &                   &60.1  & 82.3  \\
			& ReCU\cite{xu2021recu}                         &                       & 61.0 & 82.6 \\
			& \textbf{AdaBin (Ours)}                          &                   & \textbf{63.1}  & \textbf{84.3} \\
			\cline{2-5}
			& IR-Net*~\cite{qin2020forward}                 &  \multirow{4}{*}{1/1}    & 61.8 & 83.4 \\
			& Real2Bin~\cite{martinez2020training}                                          &                      & 65.4  & 86.2 \\
			& ReActNet*~\cite{liu2020reactnet}                                          &                      & 65.5  & 86.1 \\
			& \textbf{AdaBin* (Ours)}                          &                    & \textbf{66.4}  & \textbf{86.5} \\
			\midrule[1pt]
			\multirow{9}{*}{ResNet-34} & Full-precision     &         32/32       & 73.3  & 91.3     \\
			\cline{2-5}
			& ABC-Net~\cite{lin2017towards}                  & \multirow{7}{*}{1/1}  & 52.4  & 76.5     \\
			& Bi-Real~\cite{liu2018bi}                  &                    & 62.2  & 83.9  \\
			& IR-Net~\cite{qin2020forward}                   &                     & 62.9  & 84.1  \\
			& SiBNN~\cite{wang2020sparsity}                    &                     & 63.3  & 84.4  \\
			& RBNN~\cite{lin2020rotated}                     &                      & 63.1  & 84.4  \\
			& ReCU\cite{xu2021recu}                         &                        & 65.1 & 85.8 \\
			& \textbf{AdaBin (Ours)}            &                    &  \textbf{66.4}    &   \textbf{86.6}   \\
			\bottomrule[1pt]
		\end{tabular}
	\end{center}
	\caption{Comparison with state-of-the-art methods on ImageNet for AlexNet and ResNets. W/A denotes the bit width of weights and activations. * means using the two-step training setting as ReActNet.}
	\label{imagenet_results}
\end{table}

\noindent \textbf{Common structures.}\quad We show the ImageNet performance of AlexNet, ResNet-18 and ResNet-34 on Table~\ref{imagenet_results}, and compare AdaBin with recent methods like Bi-Real~\cite{liu2018bi}, IR-Net~\cite{qin2020forward}, SiBNN~\cite{wang2020sparsity}, RBNN~\cite{lin2020rotated}, ReCU\cite{xu2021recu}, Real2Bin~\cite{martinez2020training} and ReActNet~\cite{liu2020reactnet}. For AlexNet, AdaBin could greatly improve its performance on ImageNet, outperforming the current best method SiBNN by 3.4\%, and reducing the accuracy gap between BNNs and floating-point model to only 2.7\%.
Besides, AdaBin obtains a 63.1\% Top-1 accuracy with ResNet-18 structure, which only replaces the binary function and non-linear module of IR-Net~\cite{qin2020forward} with the adaptive quantizer and Maxout but gets 5.0\% improvement and outperforms the current best method ReCU by 2.1\%. For ResNet-34, AdaBin obtain 1.3\% performance improvement compared to the ReCU while binarizing the weights and activations into 1-bits. Besides, we also conduct experiments on ResNet-18 following the training setting as ReActNet. With the two step training strategy, AdaBin could get 66.4\% top-1 accuracy, which obtains 0.9\% improvement compared to ReActNet. 

\begin{table}[t]
	\setlength{\tabcolsep}{3mm}
	\renewcommand\arraystretch{0.9}
	\begin{center}
		\begin{tabular}{llcl}
			\toprule[1pt]
			Networks          		  & Methods &    OPs ($\times10^8$)  & Top-1 (\%)  \\
			\midrule[1pt]
			\multirow{2}{*}{BDenseNet28~\cite{bethge2019binarydensenet}}    & Origin  &  2.09  & 62.6   \\
			& \textbf{AdaBin}   &  2.11  & \textbf{63.7} \gbf{+1.1} \\
			\midrule[1pt]
			\multirow{2}{*}{MeliusNet22~\cite{bethge2021meliusnet}}    & Origin  &  2.08  & 63.6   \\
			& \textbf{AdaBin}   & 2.10   &  \textbf{64.6} \gbf{+1.0} \\
			\hdashline
			\multirow{2}{*}{MeliusNet29~\cite{bethge2021meliusnet}}    & Origin  &  2.14  & 65.8   \\
			& \textbf{AdaBin}   &  2.17  &  \textbf{66.5} \gbf{+0.7}  \\
			\hdashline
			\multirow{2}{*}{MeliusNet42~\cite{bethge2021meliusnet}}    & Origin  &  3.25  & 69.2   \\
			& \textbf{AdaBin}   &  3.28  &  \textbf{69.7} \gbf{+0.5}  \\
			\hdashline
			\multirow{2}{*}{MeliusNet59~\cite{bethge2021meliusnet}}    & Origin  &  5.25  & 71.0   \\
			& \textbf{AdaBin}   &  5.27  & \textbf{71.6} \gbf{+0.6} \\
			\midrule[1pt]
			\multirow{2}{*}{ReActNet-A~\cite{liu2020reactnet}}    & Origin  &  0.87  &  69.4  \\
			& \textbf{AdaBin}  &  0.88  & \textbf{70.4} \gbf{+1.0}  \\
			\bottomrule[1pt]
		\end{tabular}
	\end{center}
	\caption{Comparisons on ImageNet for binary-specific structures.}
	\label{different_structure}
\end{table}

\noindent \textbf{Binary-specific structures.}\quad Table~\ref{different_structure} shows the performance comparison with BDenseNet, MeliusNet and ReActNet. For BDenseNet28, AdaBin could get 1.1\% improvement with the same training setting, which costs negligible extra computational operations. Similarly, when AdaBin is applied to MeliusNet, an advanced version of BDenseNet, it outperforms the original networks by 1.0\%, 0.7\%, 0.5\% and 0.6\%, respectively, demonstrating that AdaBin could significantly improve the capacity and quality of binary networks. Besides, we also train the ReActNet-A structure with our AdaBin, following the same training setting with ReActNet~\cite{liu2020reactnet}. As we can see that, AdaBin could get 1.0\% performance improvement with the similar computational operations.
Our method could explicitly improve the accuracy of BNNs with a little overhead compared to state-of-the-art methods, as shown in Fig.~\ref{Acc_FLOPs}.

\begin{table}
	\setlength{\tabcolsep}{3mm}
	\renewcommand\arraystretch{0.9}
	\begin{center}
		\begin{tabular}{lcccc}
			\toprule[1pt]
			Methods          & W/A      & \#Params(M) & FLOPs(M)  & mAP  \\
			\midrule[1pt]
			Full-precision              & 32/32    & 100.28     & 31750     & 72.4 \\
			TWN~\cite{li2016ternary}             & 2/32     & 24.54      & 8531      & 67.8 \\
			DoReFa~\cite{zhou2016dorefa}      & 4/4      & 29.58      & 4661      & 69.2 \\
			\midrule[1pt]
			BNN~\cite{courbariaux2016binarized}             & 1/1      & 22.06      & 1275      & 42.0 \\
			XNOR-Net~\cite{rastegari2016xnor}            & 1/1      & 22.16      & 1279      & 50.2 \\
			BiDet~\cite{wang2020bidet}           & 1/1      & 22.06      & 1275      & 52.4 \\
			AutoBiDet~\cite{9319565}               & 1/1      & 22.06      & 1275      & 53.5 \\
			\textbf{AdaBin (Ours)}            & 1/1      & 22.47      & 1280      & \textbf{64.0}  \\
			\midrule[1pt]
			BiReal-Net~\cite{liu2018bi}      & 1/1      & 21.88      & 1277      & 63.8 \\
			BiDet*~\cite{wang2020bidet}       & 1/1      & 21.88      & 1277      & 66.0  \\
			AutoBiDet*~\cite{9319565}       & 1/1      & 21.88      & 1277      & 67.5  \\
			\textbf{AdaBin* (Ours)}        & 1/1      & 22.47      & 1282      & \textbf{69.4}   \\
			\bottomrule[1pt]
		\end{tabular}
	\end{center}
	\caption{The comparison of different methods on PASCAL VOC for object detection. W/A denotes the bit width of weights and activations. * means the the proposed method with extra shortcut for the architectures~\cite{wang2020bidet}.}
	\label{voc_results}
\end{table}

\subsection{Results on PASCAL VOC}

Table~\ref{voc_results} presents the results of object detection on PASCAL VOC dataset for different binary methods.  We follow the training strategy as BiDet~\cite{liu2018bi}. The backbone network was pre-trained on ImageNet~\cite{5206848} and then we finetune the whole network for the object detection task. During training, we used the data augmentation techniques in~\cite{wang2020bidet}, and the Adam optimizer~\cite{kingma2014adam} was applied. The learning rate started from 0.001 and decayed twice by multiplying 0.1 at the 160-th and 180-th epoch out of 200 epochs. Following the setting of BiDet~\cite{wang2020bidet}, we evaluate our proposed AdaBin on both the normal structure and the structure with real-valued shortcut. We compare them with  general binary methods BNN~\cite{courbariaux2016binarized}, XNOR-Net~\cite{rastegari2016xnor} and BiReal-Net~\cite{liu2018bi}, and also compare with BiDet~\cite{wang2020bidet} and AutoBiDet~\cite{9319565}, which are specifically designed for high-performance binary detectors. And for reference, we also show the results of the multi-bit quantization method TWN~\cite{li2016ternary} and DoReFa~\cite{zhou2016dorefa} with 4 bit weights and activations. Compared with the previous general BNNs, the proposed AdaBin improves the BNN by 22.0 mAP, XNOR by 13.8 mAP and Bi-Real Net by 5.6 mAP. Even for the task-specific optimization method BiDet, they are 11.6 mAP and 2.6 mAP lower than our method with two structures, and the improved AutoBiDet still lower than AdaBin by 10.5 mAP and 1.9 mAP. Besides, AdaBin with shortcut structure could outperform TWN and DoReFa, which demonstrates that our could significantly enable the binary neural networks to complex tasks.
\subsection{Ablation Studies}
\noindent \textbf{Effect of AdaBin quantizer.} \quad
We conduct the experiments by starting with a vanilla binary neural networks, and then add the AdaBin quantizer of weights and activations gradually. The results are shown in Table~\ref{ablation_results}, we can see that when combined with existing activation binarization by sign function, our equalization method for binarizing weights could get 0.6\% accuracy improvement. Besides, when we free the $\alpha_w$ and $\beta_w$ to two learnable parameters which are trained in an end-to-end manner as activation, it only get $86.7\%$ accuracy and is much poorer than AdaBin (the last two row). We find that its Kullback-Leibler divergence is also less than AdaBin, which shows the KLD is much important to 1-bit quantization.
When keeping the weight binarization as XNOR-Net~\cite{rastegari2016xnor}, the proposed gradient-based optimization for binarizing activations could get 1.6\% accuracy improvement, as shown in the $3^{rd}$ row. Combining the proposed weight equalization and activation optimization of AdaBin boosts the accuracy by 2\% over vanilla BNN (the $1^{st}$ vs. $4^{th}$ row), which shows that AdaBin quantizer could significantly improve the capacity of BNNs.

\noindent \textbf{Effect of $\gamma$ in Maxout.} \quad
In addition, we evaluate four activation functions on ImageNet. The first is none, denoting it is an identity connection. The second is Maxout that only with $\gamma^+$ for positive part, the third is Maxout only with $\gamma^-$ for negative part and the last one is the complete Maxout as Eq.~\ref{equ11}. As shown in Table~\ref{MaxOut_ImageNet}, the coefficient of $\gamma^+$ and $\gamma^-$ improve the accuracy by 9.6\% and 9.7\% individually. The activation function with both coefficients gets the best performance, which justifies the effectiveness of Maxout.
\begin{table}[t]
	\begin{subtable}[h]{0.58\textwidth}
		\renewcommand\arraystretch{0.8}
		\begin{center}
		\begin{tabular}{cccc}
		\toprule[1pt]
		W\_set                         & A\_set                & Non-linearity          & Acc.(\%) \\
		\midrule[1pt]
		$\{-\alpha, +\alpha\}$                 & $$\{-1, +1\}$$           & PReLU      & 85.7    \\
		$\{\textrm{w}_{b1},\textrm{w}_{b2}\}$   & $$\{-1, +1\}$$           & PReLU      & 86.3   \\
		$\{-\alpha, +\alpha\}$                  & $\{a_{b1}, a_{b2}\}$     & PReLU      & 87.3    \\
		$\{\textrm{w}_{b1},\textrm{w}_{b2}\}$   & $\{a_{b1}, a_{b2}\}$     & PReLU      & 87.7    \\
		$\{\textrm{w}_{b1},\textrm{w}_{b2}\}^*$   & $\{a_{b1}, a_{b2}\}$     &Maxout      & 86.7    \\
		$\{\textrm{w}_{b1},\textrm{w}_{b2}\}$   & $\{a_{b1}, a_{b2}\}$     &Maxout      & \textbf{88.2}    \\
		\bottomrule[1pt]
		\end{tabular}
		\end{center}
		\caption{Binary quantizer}
		\label{ablation_results}	
	\end{subtable}
	\begin{subtable}[h]{0.38\textwidth}
	\renewcommand\arraystretch{0.8}
	\begin{center}
		\begin{tabular}{ccc}
			\toprule[1pt]
			Scale factors & Top-1 (\%) & Top-5 (\%) \\
			\midrule[1pt]
			None & 53.2 & 77.2\\
			$\gamma^{+}$ & 62.8 & 83.9 \\
			$\gamma^{-}$ & 62.9 & 84.1 \\
			$\gamma^{-}$, $\gamma^{+}$ & \textbf{63.1} & \textbf{84.3} \\
			\bottomrule[1pt]
		\end{tabular}
	\end{center}
	\caption{$\gamma$ in Maxout}
	\label{MaxOut_ImageNet}
	\end{subtable}
	\caption{(a) Ablation studies of AdaBin for ResNet-20 on CIFAR-10. * means the $\alpha_w$ and $\beta_w$ are learnable parameters to the binary sets. (b) The ablation studies of Maxout on ImageNet, the scale factor with $\gamma^-$ equals to PReLU.}
\end{table}

\section{Conclusion}
In this paper, we propose an adaptive binary method (AdaBin) to binarize weights and activations with optimal value sets, which is the first attempt to relax the constraints of the fixed binary set in prior methods. The proposed AdaBin could make the binary weights best match the real-valued weights and obtain more informative binary activations to enhance the capacity of binary networks. We demonstrate that our method could also be accelerated by XNOR and BitCount operations, achieving $60.85\times$ acceleration and $31\times$ memory saving in theory. Extensive experiments on CIFAR-10 and ImageNet show the superiority of our proposed AdaBin, which outperforms  state-of-the-art methods on various architectures, and significantly reduce the performance gap between binary neural networks and real-valued networks. We also present extensive experiments for object detection, which demonstrates that our method can naturally be extended to more complex vision tasks.\\

\noindent\textit{\textbf{Acknowledgments}} This work was supported in part by Key-Area Research and Development Program of Guangdong Province No.2019B010153003, Key Research and Development Program of Shaanxi No.2022ZDLGY01-08, and Fundamental Research Funds for the Xi'an Jiaotong University No.xhj032021005-05. We gratefully acknowledge the support of MindSpore, CANN(Compute Architecture for Neural Networks) and Ascend AI Processor used for this research.

%\clearpage
\bibliographystyle{splncs04}
\bibliography{egbib}
\end{document}